\documentclass{llncs}

\makeatletter 
\renewcommand\@biblabel[1]{#1.} 
\makeatother

\usepackage[]{algorithm2e} 
\usepackage[dvipsnames]{xcolor}
\usepackage{amsmath}
\usepackage{amssymb}
\usepackage{booktabs}
\usepackage{color}
\usepackage{graphicx}

\usepackage[numbers, sort&compress]{natbib}
\usepackage{tabularx}

\begin{document}
\mainmatter 

\title{Shallow reading with Deep Learning: \protect\\ Predicting popularity of online content\\ using only its title}
\titlerunning{Shallow reading with Deep Learning}

\author{Wojciech Stokowiec\inst{1, \, 3}, Tomasz Trzci\'{n}ski\inst{2, \, 3}, Krzysztof Wo\l{}k\inst{1},\\ Krzysztof Marasek\inst{1}  and Przemys\l{}aw Rokita\inst{2}}
\authorrunning{Wojciech Stokowiec et al.} 
%
\tocauthor{Wojciech Stokowiec et al.}
\institute{Polish-Japanese Academy of Information Technology, Warsaw, Poland,\\
\email{wojciech.stokowiec@pjwstk.edu.pl},
\and
Warsaw University of Technology, Warsaw, Poland \\
\email{t.trzcinski@ii.pw.edu.pl}, \email{pro@ii.pw.edu.pl} \\
\and 
Tooploox, Poland\\
}

\maketitle

\begin{abstract}
With the ever decreasing attention span of contemporary Internet users, the title of online content (such as a news article or video) can be a major factor in determining its popularity. To take advantage of this phenomenon, we propose a new method based on a bidirectional Long Short-Term Memory (LSTM) neural network designed to predict the popularity of online content using only its title. We evaluate the proposed architecture on two distinct datasets of news articles and news videos distributed in social media that contain over 40,000 samples in total. On those datasets, our approach improves the performance over traditional shallow approaches by a margin of 15\%. Additionally, we show that using pre-trained word vectors in the embedding layer improves the results of LSTM models, especially when the training set is small.   
To our knowledge, this is the first attempt of applying popularity prediction using only textual information from the title.
\end{abstract}
\section{Introduction}

The distribution of textual content is typically very fast and catches user attention for only a short period of time~\citep{Castillo14}. For this reason, proper wording of the article title may play a significant role in determining the future popularity of the article. The reflection of this phenomenon is the proliferation of {\it click-baits} - short snippets of text whose main purpose is to encourage viewers to click on the link embedded in the snippet. Although detection of click-baits is a separate research topic~\citep{Chakraborty16}, in this paper we address a more general problem of predicting popularity of online content based solely on its title.

Predicting popularity in the Internet is a challenging and non-trivial task due to a multitude of factors impacting the distribution of the information: external context, social network of the publishing party, relevance of the video to the final user, etc. This topic has therefore attracted a lot of attention from the research community~\citep{Szabo10,Pinto13,Castillo14,Ramisa16}.

In this paper we propose a method for online content popularity prediction based on a bidirectional recurrent neural network called BiLSTM. This work is inspired by recent successful applications of deep neural networks in many natural language processing problems~\citep{Collobert2011,ZhangL15}. Our method attempts to model complex relationships between the title of an article and its popularity using novel deep network architecture that, in contrast to the previous approaches, gives highly interpretable results. Last but not least, the proposed BiLSTM method provides a significant performance boost in terms of prediction accuracy over the standard shallow approach, while outperforming the current state-of-the-art on two distinct datasets with over 40,000 samples.

To summarize, the contributions presented in this paper are the following:
\begin{itemize}
\item Firstly, we propose title-based method for popularity prediction of news articles  based on a deep bidirectional LSTM network.
\item Secondly, we show that using pre-trained word vectors in the embedding layer improves the results of LSTM models.
\item Lastly, we evaluate our method on two distinct datasets and show that it outperforms the traditional shallow approaches by a large margin of 15\%.
\end{itemize}

The remainder of this paper is organized in the following manner: first, we review the relevant literature and compare our approach to existing work. Next, we formulate the problem of popularity prediction and propose a  model that takes advantage of BiLSTM architecture to address it. Then, we evaluate our model on two datasets using several pre-trained word embeddings and compare it to benchmark models. We conclude this work with discussion on future research paths. 

\section{Related Work}

The ever increasing popularity of the Internet as a virtual space to share content inspired research community to analyze different aspects of online information distribution. Various types of content were analyzed, ranging from textual data, such as Twitter posts~\citep{Castillo14} or Digg stories~\citep{Szabo10} to images~\citep{Khosla14} to videos~\citep{Chesire01,Pinto13, TrzcinskiR15}. Although several similarities were observed across content domains, e.g. log-normal distribution of data popularity~\citep{Tsagkias10}, in this work we focus only on textual content and, more precisely, on the popularity of news articles and its relation to the article's title.

Forecasting popularity of news articles was especially well studied in the context of Twitter - a social media platform designed specifically for sharing textual data~\citep{Bandari12,Hong11}. Not only did the previous works focus on the prediction part, but also on modeling message propagation within the network~\citep{Osborne11}. However, most of the works were focused on analyzing the social interactions between the users and the characteristics of so-called social graph of users' connections, rather than on the textual features. Contrary to those approaches, in this paper we base our predictions using only textual features of the article title. We also validate our proposed method on one dataset collected using a different social media platform, namely Facebook, and another one created from various news articles~\citep{Ramisa16}.

Recently, several works have touched on the topic of popularity prediction of news article from a multimodal perspective~\citep{Ramisa16,Chen16}. Although in~\citep{Ramisa16} the authors analyze news articles on a per-modality basis, they do not approach the problem of popularity prediction in a holistic way. To address this shortcoming, \citep{Chen16} have proposed a multimodal approach to predicting popularity of short videos shares in social media platform Vine\footnote{\url{https://vine.co/}} using a model that fuses features related to different modalities. In our work, we focus only on textual features of the article title for the purpose of popularity prediction, as our goal is to empower the journalists to quantitatively assess the quality of the headlines they create before the publication. Nevertheless, we believe that in future research we will extend our method towards multimodal popularity prediction.
 
\section{Method}

In this section we present the bidirectional LSTM model for popularity prediction. We start by formulating the problem and follow up with the description of word embeddings used in our approach. We then present the Long Short-Term Memory network that serves as a backbone for our bidirectional LSTM architecture. We conclude this section with our interpretation of hidden bidirectional states and describe how they can be employed for title introspection. 

\subsection{Problem Formulation}
We cast the problem of popularity prediction as a binary classification task. We assume our data points contain a string of characters representing article title and a popularity metric, such as number of comments or views. The input of our classification is the character string, while the output is the binary label corresponding to popular or unpopular class. To enable the comparison of the methods on datasets containing content published on different websites and with different audience sizes, we determine that a video is popular if its popularity metric exceeds the median value of the corresponding metric for other points in the set, otherwise - it is labeled as unpopular. The details of the labeling procedure are discussed separately in the Datasets section.

\subsection{Text Representation}
Since the input of our method is textual data, we follow the approach of~\citep{Salton1975} and map the text into a fixed-size vector representation. To this end, we use word embeddings that were successfully applied in other domains. We follow \cite{Collobert2011} and use pre-trained GloVe word vectors~\citep{glove} to initialize the embedding layer (also known as look-up table). Section \ref{seq:embeddings} discusses the embedding layer in more details. 

\subsection{Bidirectional Long Short-Term Memory Network}
\begin{figure}[h!]
\centering
\includegraphics[height=7cm]{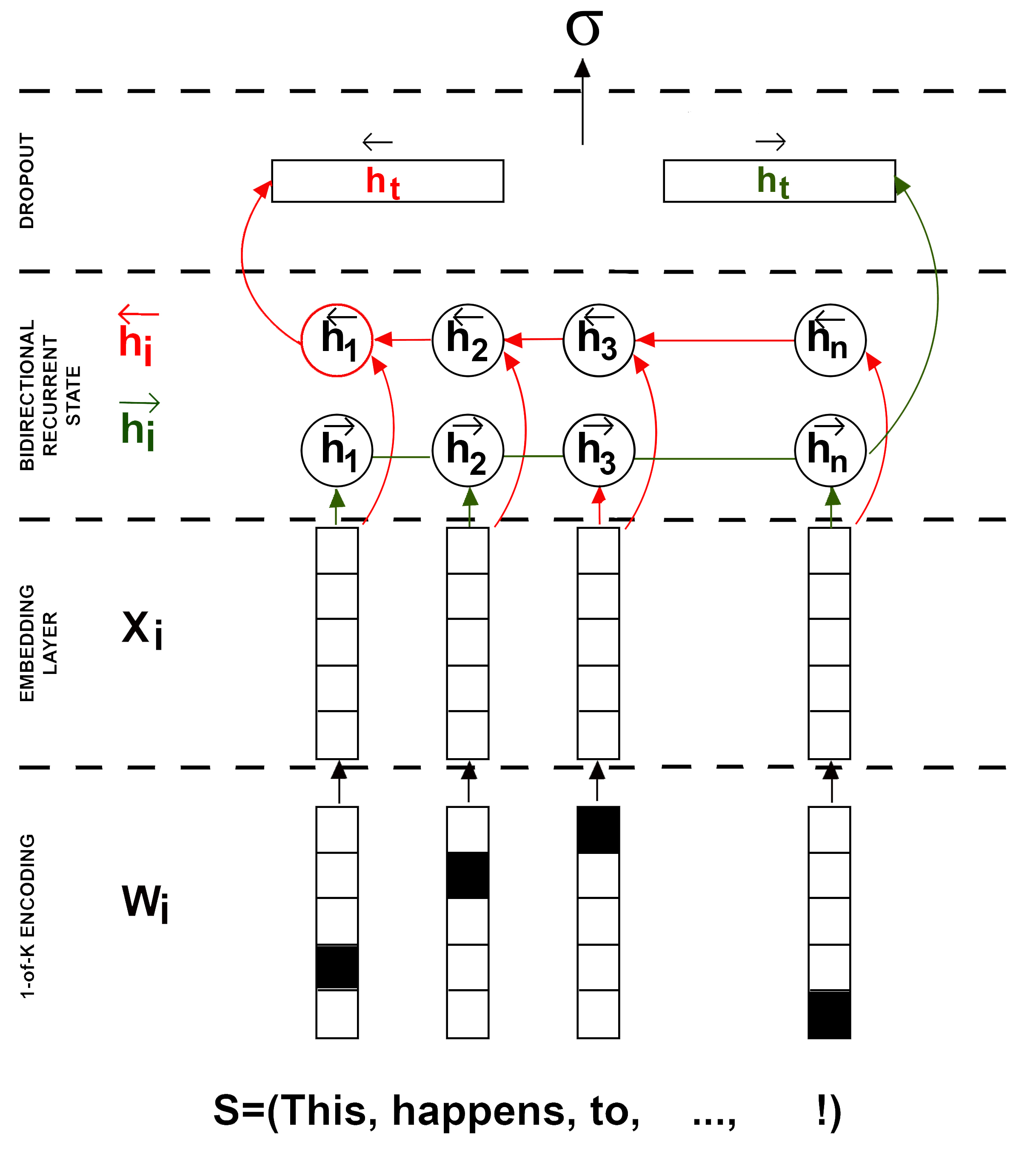}
\caption{A bidirectional LSTM architecture with $1$-of-$K$ word encoding and embedding layer proposed in this paper.}
\label{biLSTM}
\end{figure}

Our method for popularity prediction using article's title is inspired by a bidirectional LSTM architecture. The overview of the model can be seen in Fig.~\ref{biLSTM}.

Let $x_i \in \mathbb{R}^d$ be $d$-dimensional word vector corresponding to the $i$-the word in the headline, then a variable length sequence:  $\mathbf{x}  = (x_1,  x_2, \ldots, x_n )$ represents a headline.  
A recurrent neural network  (RNN) processes this sequence by recursively applying a transformation function to the current element of sequence $x_t$ and its previous hidden internal state $h_{t-1}$ (optionally outputting $y_t$). At each time step $t$, the hidden state is updated by:
\begin{equation}
	h_t  = \sigma(W_h \cdot [h_{t-1}, x_t] + b_h), 
\end{equation}
where $\sigma$ is a non-linear activation function. LSTM network~\citep{LSTM} updates its internal state differently, at each step $t$ it calculates:
\begin{align}
\begin{split}
i_t &= \sigma(W_i \cdot [h_{t-1}, x_t] + b_i ),\\  
f_t &= \sigma(W_f \cdot [h_{t-1}, x_t] + b_f ),\\  
o_t &= \sigma(W_o \cdot [h_{t-1}, x_t] + b_o ), \\ 
\widetilde{c_t} &= \tanh(W_{\widetilde{c}} \cdot [h_{t-1}, x_t] + b_{\widetilde{c}}), \\ 
c_t &= f_t \odot c_{t-1} + i_t \odot \widetilde{c_t}, \\
h_t &= o_t \odot \tanh(c_t),
\label{lstm}
\end{split}
\end{align}
where $\sigma$ is the \textit{sigmoid} activation function, \textit{tanh} is the hyperbolic tangent function and $\odot$ denotes component-wise multiplication.  In our experiments we used 128, 256 for the dimensionality of hidden layer in both LSTM and BiLSTM. The~term in equation \ref{lstm} $i_t$, is called the input gate and it uses the input word and the past hidden state to determine whether the input is worth remembering or not. The amount of information that is being discarded is controlled by forget gate $f_t$ , while $o_t$ is the output gate that controls the amount of information that leaks from memory cell $c_t$ to the hidden state $h_t$. 
In the context of classification, we typically treat the output of the hidden state at the last time step of LSTM as the document representation and feed it to sigmoid layer to perform classification~\citep{ZhouSLL15b}. 

Due to its sequential nature, a recurrent neural network puts more emphasis on the recent elements. To circumvent this problem \citep{BiLSTM} introduced a bidirectional RNN in which each training sequence is presented forwards and backwards to two separate recurrent nets, both of which are connected to the same output layer. Therefore, at any time-step we have the whole information about the sequence. This is shown by the following equation:
\begin{align}
\begin{split}
	\overrightarrow{h_t}  & =  \sigma(\overrightarrow{W_h} \cdot [h_{t-1}, x_t] +\overrightarrow{b_h}), \\
 	\overleftarrow{h_t}  & =  \sigma(\overleftarrow{W_h} \cdot [h_{t+1}, x_t] +\overleftarrow{b_h}).  \\
\end{split}
\end{align}

In our method, we use the bidirectional LSTM architecture for content popularity prediction using only textual cues. We have to therefore map the neural network outputs from a set of hidden states $(\overrightarrow{h_i}, \overleftarrow{h_i})_{i \in 1 \ldots n}$ to classification labels. We evaluated several approaches to this problem, such as max or mean pooling. The initial experiments showed that the highest performance was achieved using late fusion approach, that is by concatenating the last hidden state in forward and backward sequence. The intuition behind this design choice is that the importance of the first few words of the headline is relatively high, as the information contained in $\overleftarrow{h_1}$, i.e. the last item in the backward sequence, is mostly taken from the first word.

\subsection{Hidden State Interpretation}

One interesting  property of bidirectional RNNs is the fact, that the concatenation of hidden states $\overrightarrow{h_t} $ and $\overleftarrow{h_t}$ can be  interpreted  as a context-dependent vector representation of word $w_t$ . This allows us to introspect a given title and approximate the contribution of each word to the estimated popularity.  
	To that end one can process the  headline representation $\mathbf{x}  = (x_1,  x_2, \ldots, x_n )$  through the bidirectional recurrent network and then retrieve pairs of forward and backwards hidden state $[ \overrightarrow{h_t}, \overleftarrow{h_t} ] $ for each word $x_t$.  Then, the output of the last fully-connected layer  $\sigma([ \overrightarrow{h_t}, \overleftarrow{h_t} ])$ could be interpreted as context-depended popularity of a word $x_t$. 
    
\subsection{Training}
In our experiments we minimize the binary cross-entropy loss using Stochastic Gradient Descent on randomly shuffled mini-batches with the Adam optimization algorithm \citep{ADAM}. We reduce the learning rate by a factor of 0.2 once learning plateaus. We also employ early stopping strategy, i.e. stopping the training algorithm before convergence based on the values of loss function on the validation set. 
 
\section{Evaluation}

In this section, we evaluate our method and compare its performance against the competitive approaches. We use $k$-fold evaluation protocol with $k=5$ with random dataset split. We measure the performance using standard accuracy metric which we define as a ratio between correctly classified data samples from test dataset and all test samples. 

\subsection{Datasets\label{datasets}}
In this section we present two datasets used in our experiments: The NowThisNews dataset, collected for the purpose of this paper, and The BreakingNews dataset~\citep{Ramisa16}, publicly available dataset of news articles.

\subsubsection*{The NowThisNews Dataset} contains  4090 posts with associated videos from NowThisNews Facebook page\footnote{\url{https://www.facebook.com/NowThisNews}} collected between 07/2015 and 07/2016. For each post we collected its title and the number of views of the corresponding video, which we consider our popularity metric. Due to a fairly lengthy data collection process, we decided to normalize our data by first grouping posts according to their publication month and  then labeling the posts for which the popularity metric exceeds the median monthly value as popular, the remaining part as unpopular. 

\subsubsection*{The Breaking News Dataset} \citep{Ramisa16} contains a variety of news-related information such as images, captions, geo-location information and comments which could be used as a proxy for article popularity.  The articles in this dataset were collected between January and December 2014. Although we tried to retrieve the entire dataset, we were able to download only 38,182 articles due to the dead links published in the dataset. The retrieved articles were published in main news channels, such as Yahoo News, The Guardian or The Washington Post. Similarly, to The NowThisNews dataset we normalize the data by grouping articles per publisher, and classifying them as popular, when the number of comments exceeds the median value for given publisher. 

\subsection{Baselines}

As a first baseline we use Bag-of-Words, a well-known and robust text representations used in various domains~\citep{WangManning}, combined with a standard shallow classifier, namely, a Support Vector Machine with linear kernel. We used LIBSVM\footnote{https://www.csie.ntu.edu.tw/~cjlin/libsvm/} implementation of  SVM.

Our second baseline is a deep Convectional Neural Network applied on word embeddings. This baseline represents state-of-the-art method presented in~\citep{Ramisa16} with minor adjustments to the binary classification task.
	The architecture of the CNN benchmark we use is the following: the embedding layer transforms one-hot encoded words to their dense vector representations, followed by the convolution layer of 256 filters with width equal to 5 followed by max pooling layer (repeated three times), fully-connected layer with dropout and $l_2$ regularization and finally, \textit{sigmoid} activation layer. For fair comparison, both baselines were trained using the same training procedure as our method.

\subsection{Embeddings}\label{seq:embeddings}
As a text embedding in our experiments, we use publicly available GloVe word vectors~\citep{glove} pre-trained on two datasets: Wikipedia 2014 with Gigaword5 (W+G5) and Common Crawl (CC)\footnote{\url{http://nlp.stanford.edu/projects/glove/}}. Since their output dimensionality can be modified, we show the results for varying dimensionality sizes. On top of that, we evaluate two training approaches: using static word vectors and fine-tuning them during training phase. 

\subsection{Results}
The results of our experiments can be seen in Tab.~\ref{results_nowthisnews} and~\ref{results_breakingnews}. Our proposed BiLSTM approach outperforms the competing methods consistently across both datasets. The performance improvement is especially visible for The NowThisNews dataset and reaches over 15\% with respect to the shallow architecture in terms of of accuracy. Although the improvement with respect to the other methods based on deep neural network is less evident, the recurrent nature of our method provides much more intuitive interpretation of the results and allow for parsing the contribution of each single word to the overall score.

%
%
\begin{table}[ht]
\setlength{\tabcolsep}{4pt}
\def\arraystretch{0.95}
\centering
\caption{Popularity prediction results on NowThisNews dataset. Our proposed BiLSTM method provides higher performances than the competitors in terms of classification accuracy.}
\label{results_nowthisnews}
\begin{tabular}{@{}lllllll@{}}
\toprule
Model 		& Word Embeddings  & fine-tuned       & Dim  	& Accuracy 	 	\\ \midrule
BoW + SVM 	& 				   & 				  &         & 0.5832     	\\ \midrule
CNN   		& GloVe (W + G5) & no 				& 100    	& 0.6320     	\\
   	  	 	& GloVe (W + G5) & yes 				& 100    	& 0.6454     	\\
            & GloVe (W + G5) & no 				& 200    	& 0.6308     	\\
            & GloVe (W + G5) & yes 				& 200    	& 0.6479     	\\

            & GloVe (W + G5) & no 				& 300 	 	& 0.6247     	\\
            & GloVe (W + G5) & yes 				& 300    	& 0.6295     	\\
            & GloVe (CC)     & no 				& 300   	& 0.6528     	\\
            & GloVe (CC)	 & yes  			& 300   	&  { 0.6653} 	\\ \midrule
      
LSTM 128 	& Glove (W + G5) & no 			    & 300 & 0.63081  			\\ 
			& Glove (W + G5) & yes 				& 300 & 0.64792  			\\ 
LSTM 256 	& Glove (W + G5) & no 			    & 300 & 0.64914  			\\ 
		 	& Glove (W + G5) & yes 				& 300 & 0.66504   			\\ \midrule

BiLSTM 128 	& Glove (W + G5) & no 			    & 300 & 0.6552 	  			\\ 
			& Glove (W + G5) & yes 			    & 300 & 0.6479 	  			\\ 
BiLSTM 256 	& Glove (W + G5) & no 			    & 300 & 0.6564 	  			\\ 
			& Glove (W + G5) & yes 			    & 300 & \textbf{0.6711}  	\\
\bottomrule
\end{tabular}
\end{table}

%
%
\begin{table}
\setlength{\tabcolsep}{6pt}
\def\arraystretch{0.95}
\centering
\caption{Popularity prediction results on BreakingNews dataset. Our BiLSTM method outperforms the competitors - the performance gain is especially visible with respect to the shallow architecture of BoW + SVM.}
\label{results_breakingnews}
\begin{tabular}{@{}lllllll@{}}
\toprule
Model 		& Word Embeddings  & fine-tuned       & Dim  & Accuracy 	 \\ \midrule
BoW + SVM 	& 				   & 				  &           & 0.7300   \\ \midrule
CNN   		& GloVe (W + G5) & no 				& 100    	& 0.7353     \\
			& GloVe (W + G5) & yes 				& 100    	& 0.7412     \\
            & GloVe (W + G5) & no 				& 200    	& 0.7391     \\
            & GloVe (W + G5) & yes 				& 200    	& 0.7379     \\
            & GloVe (W + G5) & no 				& 300    	& 0.7319     \\
            & GloVe (W + G5) & yes 				& 300    	& {0.7416}   \\
            & GloVe (CC)     & no 				& 300   	& 0.7355     \\         
            & GloVe (CC)	 & yes  			& 300   	& 0.7394     \\\midrule
      
LSTM 128 	& Glove (W + G5) & yes 				& 300  & 0.6694 		 \\ 
			& Glove (W + G5) & no 			    & 300  & 0.6663 		 \\ 
LSTM 256 	& Glove (W + G5) & yes 				& 300  & 0.6619 		 \\ 
		 	& Glove (W + G5) & no 			    & 300  & 0.6624 		 \\ \midrule

BiLSTM 128 	& Glove (W + G5) & yes 			    & 300  & 0.7167 		 \\ 
			& Glove (W + G5) & no 			    & 300  & 0.7406 		 \\ 
BiLSTM 256 	& Glove (W + G5) & yes 			    & 300  & 0.7149 		 \\ 
			& Glove (W + G5) & no 			    & 300  & \textbf{0.7450} \\ 

\bottomrule
\end{tabular}
\end{table}

To present how our model works in practice, we show in Tab.~\ref{topandbottom} a list of 3 headlines from NowThisNews dataset that are scored with the highest probability of belonging to a popular class, as well as 3 headlines with the lowest score. As can be seen, our model correctly detected videos that become viral at the same time assigning low score to content that underperformed. We believe that BiLSTM could be successfully
applied in real-life scenarios. 

\begin{table}[ht]
\def\arraystretch{1.1}
\centering
\caption{Top and bottom 3 headlines from the NowThisNews dataset as predicted by our model and their views 168 hours after publication.}
\label{topandbottom}
\begin{tabularx}{\textwidth}{Xr}
\toprule
\textbf{Top 3 headlines}                                                                      & \textbf{Views} 
\\ \midrule
This teen crossed a dangerous highway to play ‘Pokémon Go’ — and then was hit by a car           & 20'836'692     
\\
This dancer dropped her phone in the water but a dolphin had her back                            & 1'887'482   
\\ 
A man shoved a bag of sh*t down this woman’s pants — and was caught on camera                    & 784'588                       
\\ 
\toprule
\textbf{Bottom 3 headlines}																						& \textbf{Views} 
\\ \midrule
We're recapping some of the biggest stories from last night and this morning                                  	& 47'803              
\\
We're recapping some of the big stories you might have missed                                                   & 64'740   
\\
Violent clashes between protesters and police broke out in Hong Kong                                            & 256'357  
\\
\bottomrule
\end{tabularx}
\end{table}

\section{Conclusions}

In this paper we present a novel approach to the problem of online article popularity prediction. To our knowledge, this is the first attempt of predicting the performance of content on social media using only textual information from its title.  We show that our method consistently outperforms benchmark models. Additionally, the proposed method could not only be used to compare competing titles  with regard to their estimated probability, but also to gain insights about what constitutes a good title.  
Future work includes modeling popularity prediction problem with multiple data modalities, such as images or videos. Furthermore, all of the evaluated models  function at the word level,  which could be problematic due to idiosyncratic nature of social media and Internet content. It is, therefore, worth investigating, whether combining models that operate at the character level to learn and generate vector representation of titles  with visual features could improve the overall performance. 

\subsubsection*{Acknowledgment}
The authors would like to thank NowThisMedia Inc. for enabling this research by providing access to data and hardware.

\small{
	\bibliography{example}
}

\end{document}